\documentclass[letterpaper, 10 pt, conference]{ieeeconf}  

\IEEEoverridecommandlockouts                              

\usepackage{graphicx} 
\usepackage{xcolor}
\usepackage{algorithm}
\usepackage{algpseudocode}
\usepackage{caption}
\usepackage{subcaption}
\newtheorem{definition}{Definition}
\title{\LARGE \bf
A COLREGs-Compliant Conflict Resolution Strategy for Autonomous Surface Vehicles
}

\author{Raghav Thakar, Rajat Agrawal and Sujit PB
\thanks{This work is partially supported by the SERB CORE Grant CRG/2021/007916. }
\thanks{Raghav Thakar is an undergraduate student at Manipal Institute of Technology, Manipal, Karnataka. Email: raghav.thakar@learner.manipal.edu}
\thanks{Rajat Agrawal and P.B. Sujit are with the Department of Electrical Engineering and Computer Science at IISER Bhopal, Bhopal--462066, India. Email: (arajat,sujit)@iiserb.ac.in.}%
}

\begin{document}

\maketitle
\thispagestyle{empty}
\pagestyle{empty}

\begin{abstract}

This paper presents a novel conflict resolution strategy for autonomous surface vehicles (ASVs) to safely navigate and avoid collisions in a multi-vessel environment at sea. Collisions between two or more marine vessels must be avoided by following the International Regulations for Preventing Collisions at Sea (COLREGs). We propose strategy a two-phase strategy called as COLREGs Compliant Conflict-Resolving (COMCORE) strategy,   that generates collision-free trajectories for ASVs while complying with COLREGs. In phase-1, a shortest path for each agent is determined, while in phase-2 conflicts are detected and resolved by modifying the path in  compliance with COLREGs. COMCORE solution optimises vessel trajectories for lower costs while also providing a safe and collision-free plan for each vessel. Simulation results are presented to show the applicability of COMCORE for larger number agents with very low computational requirement and hence scalable. Further, we  experimentally demonstrate COMCORE for two ASVs in a lake to show its ability to determine solution and implementation capability in the real-world. 
\end{abstract}

\section{INTRODUCTION}
The applications of Autonomous Surface Vehicles (ASVs) in the maritime domain are increasing day by day. These ASVs need to operate with other vessels that are manually driven. Thus, the ship captain/helmsman of the boat anticipates that the ASV to follow the very rules that other marine vessels follow to avoid collisions. Therefore, there is a need to develop collision avoidance algorithms for ASVs that adhere to the rules prescribed in International Regulations for Preventing Collisions at Sea (COLREGs).

There are several collision avoidance approaches developed in the literature, like velocity obstacles based methods\cite{van2011reciprocal, fiorini1998motion}, potential field based methods \cite{khatib1986real, sun2017collision}, navigation function method \cite{roussos20083d},  vector field-based methods \cite{panagou2014motion}, control based frameworks \cite{seiler1998application, hu2020real}, and geometry-based frameworks \cite{chakravarthy1998obstacle}. These approaches are not directly applicable for the marine vessels as they interact with manned marine vehicles and hence are required to meet deterministic behaviours by satisfying the COLREGs rules. 

Some of the above approaches have been modified to meet the COLREGs requirement. A survey of collision avoidance approaches for ASVs is given in \cite{zhang21}. We review some of the works that are closely related to our work where we would like to develop a collision avoidance system for ASVs while satisfying COLREGs requirements. In \cite{8281087}, an RRT-based approach with global and local optimisation techniques to generate feasible solutions is used. The paths generated by RRT can be highly sub-optimal. In \cite{8463182}, a model predictive control approach is developed such that only certain half-spaces are active for the MPC control evaluation. The MPC optimizes the control and deviation from the desired trajectory and hence its performance is better than the RRT approach. However, the MPC approach can become computationally heavy and requires model of the other vehicles. In \cite{kufoalor2018proactive, thyri2022partly}, the reciprocal velocity obstacle method \cite{van2011reciprocal} is modified to meet the COLREGs requirements.  This approach is a reactive which may make the vehicle deviate more than required. 

The collision between two vessels occur when they are unable to resolve the conflict in their paths. We can think of collision avoidance as conflict-resolution, where the resolution is to find alternative paths for either a single vessel or for both the vessels. The conflict-resolution between two vessels then can be extended to multiple vessels by taking a pair of vessels at a time. In this paper, we propose to develop a collision avoidance and conflict resolution method influenced by search-based planning \cite{SHARON201540} approach but are not COLREGs complaint. 
The main contributions of this paper are (i)
    to develop a novel COLREGs compliant conflict resolution strategy  (COMCORE) for collision avoidance (ii) Evaluate the approach on larger of vessels and shows it scalability properties
    and (iii)  experimentally validate COMCORE on ASVs in a lake. 


\section{COLREGs-Compliant Collision Avoidance Rules}
The guidelines outlined by COLREGs are spread across 41 rules. In this paper, we focus on  Section II, Rules 8 and 13-17 of COLREGs, that occur the most in the maritime scenarios. These rules specify how to identify possible collision scenarios, and the actions to be taken by the involved vessels \cite{8281087}. We define the following two collision situations that occur in the waters:
\begin{definition}[{\bf Head-On}]
The vessels are meeting on a reciprocal or near-reciprocal course with a risk of collision. The vessel headings angle are opposite to each other.
\end{definition}
\begin{definition}[{\bf Crossing}] The vessels are crossing so as to involve a risk of collision. In this scenario, the heading angle of the vessels are orthogonal.
\end{definition}

\begin{figure}
    \centering
    \includegraphics[width=\linewidth]{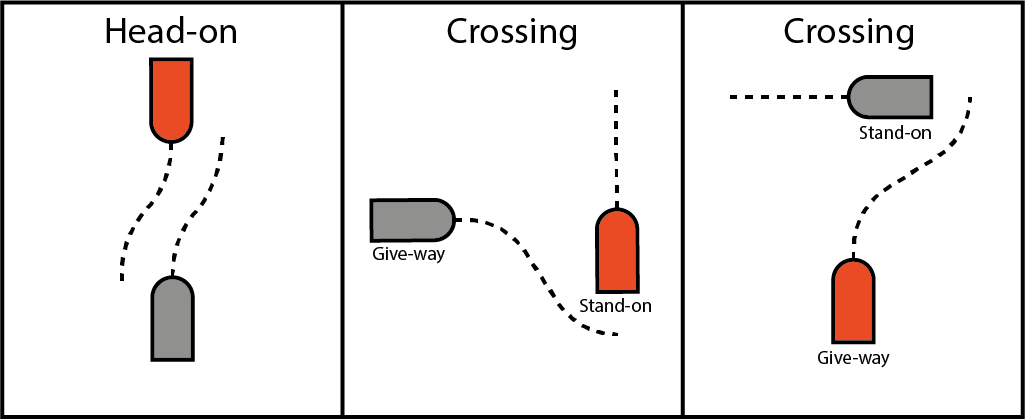}
    \caption{Collision Situations and Appropriate Vessel Behaviour by Give-way and Stand-on vessels.}
    \label{fig:COLREGs-behaviour}
\end{figure}

The action taken by both vessels is identical in Head-On situations. However, in Crossing situations, each of the two involved vessels must first be labeled as Give-way vessel or Stand-on vessel. The labels are defined as 
\begin{definition}[{\bf Give-way}] The vessel which is directed to keep out of the way of another vessel. According to Rule 15 of COLREGs, in Crossing situations, the vessel which has the other on her own starboard side is labeled as \textbf{Give-way}.
    \end{definition}
\begin{definition}[{\bf Stand-on}]The vessel that shall keep her course and speed based on Rule 15 of COLREGs.
    \end{definition}
Fig. \ref{fig:COLREGs-behaviour} depicts the collision situations considered in this paper and the COLREGs appropriate behaviour by each involved vessel.

\section{COlregs-coMpliant COnflict-REsolving (COMCORE) strategy}
\subsection{Environment}
In order to move the vessel from one point to another the path is defined in terms of waypoints. In this paper, we consider the path to be a set of waypoints. The workspace area $L\times W~ m^2$ is decomposed into a grid, where each cell is of length $\ell \times \ell$, and the total number of cell are $\mathbf{N}$. The center of the cell is called the vextex (or node) $v_m$. We restrict motion of the vessel from a cell ($v_m$, for some $m\in \mathbf{N}$) to either move forward, right or left. The neighbours of $v_m$, $\mathcal{N}_m$, are the cells that can be reached from $v_m$. An edge is the path segment connecting $v_m$ to $v_m',m'\in\mathcal{N}_m$. Using the vextices and edges, we can create a graph $G=(V,E)$. Note that the vessels move continuous and use the nodes as the waypoints. Thus, we will use the grid structure of the environment to determine the conflict resolution path. 

We assume that there are $k$ agents $a_{1}, a_{2},\ldots,a_{k}$ with their initial positions as $s_{1}, s_{2},\ldots, s_{k},s_x\in\mathbf{R}^2$, and goal positions $g_{1}, g_{2},\ldots,g_{k}$ on a discrete workspace. For describing the COMCORE stratergy, we need some additional definitions that are defined as:


 \begin{definition}[Path]
     A \textit{path} is a set of cells for a single agent $a_k$ from $s_k$ to $g_k$. 
 \end{definition}
 
 \begin{definition}[Solution]
     Solution is the set of $k$ conflict resolved paths for the $k$ agents \cite{SHARON201540}.
 \end{definition}
 %
 \begin{definition}[Conflict]\label{def:conflict}
 A \textit{conflict} is defined as a tuple $(a_{i}, a_{j}, v_m, t,ctype )$ where agent $a_{i}$ and $a_{j}$ occupy the same vertex $v_m$ at time $t$. $ctype$ refers to the type of conflict ({Head-on} or {Crossing}). 
 \end{definition}
 The $ctype$ field enables the agents to determine the feasible regions while satisfying the COLREGs. 

COMCORE uses a 2-phase decision-making architecture to resolve conflicts and determine the solution. These process of these two phases are given below.

\subsection{Phase-1: Path-Finding for Individual Agents}
The phase-1 operation of COMCORE finds optimal paths for each agent from the given start to goal positions. We use A* \cite{4082128} to determine the optimal paths. As phase-1 and phase-2 processes are independent, one can select another motion planning algorithm to determine the optimal paths instead of A*. 



The Algorithm \ref{alg:phase_1} returns a set of optimal paths for each agent, and this set of paths becomes the stating point for phase-2 operation.

\begin{algorithm}
\caption{Phase-1 COMCORE Algorithm}
\begin{algorithmic}[1]
    \State Input: $k$, $s_1,s_2,\ldots,s_k$, $g_1,g_2,\ldots,g_k$, workspace
    \State Output: Set $S$ of paths for each agent
    \State $i \gets 0$
    \While{$i \leq k$}
        \State $S[i] \gets $ A$^*$($i, g_{i}, s_{i}$)
    \EndWhile
    \State \textbf{return} $S$
\end{algorithmic}
\label{alg:phase_1}
\end{algorithm}

\subsection{Phase-2: Resolving Conflicts and Making Paths COLREGs-Compliant}

The phase-2 operation of COMCORE identifies conflicts in a solution and resolves them to make the paths COLREGs-compliant.  It does this by building and growing a \textit{solution list} (SL) that stores a solution (a set of $k$ paths) in its every node. A new node is added every time a conflict is found and resolved. The new node contains the solution with the newest conflict having been resolved. A node $N$ in SL is a goal node when the solution it stores is valid, i.e., all its paths are free from any conflicts. 

Every node $N$ in the SL has the following member data and methods to realise the above functionalities:

\begin{itemize}
    \item[] {\textbf{\textit{N.solution}}} The $k$ paths that are stored in the node.
    \item[] \textbf{\textit{N.validate()}}: To check the paths in the solution for conflicts.
    \item[] \textbf{\textit{N.resolve()}}: To modify the paths in the solution to make them COLREGs-compliant at the conflict.
\end{itemize}

\subsubsection{Phase-2 Process}
We start by initialising the SL with an empty root node $N$ and calling the phase-1 pathfinder to find paths for the $k$ agents in the system. This solution is stored in $N$. We validate its solution using the \textit{$N$.validate()} function. If the validation process does not yield any conflict, the node is labeled as a goal node and the algorithm ends. If a conflict is found, it is passed to the \textit{$N$.resolve()} function to make the paths COLREGs-compliant at the conflict. We create a child node $C$ of $N$ and initialise it with the modified solution returned by \textit{$N$.resolve()} and repeat the process by validating the new solution. Algorithm \ref{alg: high-level-algorithm} formally describes this phase-2 COMCORE algorithm.

\subsubsection{Solution Validation}
The \textit{validate()} method identifies and returns a conflict in a node's solution. It simulates the paths computed for each agent to find an instance of two agents occupying the same vertex at the same timestep. When such an instance is found, the conflict is recorded in a conflict tuple as described in Definition \ref{def:conflict}. While it is possible for more than two agents to be involved in colliding paths, the validation process records only two conflicting agents at a time. This is to allow the direct application of COLREGs during the conflict resolution process. Also, the validation process halts upon finding the first conflict in the solution. This results in each node in SL being responsible for identifying and then resolving only one conflict, making the COMCORE strategy fit for modifications to consider many unique conflict resolution techniques.

\begin{algorithm}
\caption{Phase-2 COMCORE Algorithm}
\begin{algorithmic}[1]
\State Input: Workspace, $k$ agents, $k$ start positions, $k$ goal positions, $S$ (from Algorithm \ref{alg:phase_1})
\State Output: Solution List SL
  \Procedure{Phase2}{}
  \State $N \gets$ \textbf{new} $Node$
  \State $N.solution \gets S$
  \While{$True$}
  \State $conf \gets N.validate()$ \Comment{Returns conflict}
    \If{$conf == None$}
      \State \textbf{return} $N$ \Comment{$N$ is a goal node}
    \Else
        \State $C \gets$ \textbf{new} $Node$
        \State $C.solution  \gets N.resolve(N.solution, \; conf)$
    \EndIf
    \State $N \gets C$
  \EndWhile
  \EndProcedure
\end{algorithmic}
\label{alg: high-level-algorithm}
\end{algorithm}

\subsubsection{Identifying the Conflict Type}
The $ctype$ field in a conflict tuple stores the type of collision situation that that conflict would create (Head-on or Crossing). This information is important to plan the appropriate actions by each vessel during the conflict resolution process. During validation, when a collision in paths is identified, we record the time $t$ and vertex $v_{t}$ of the collision. We then record the positions of both vessels one time step before and after the collision. We call this the \textit{collision window}. Table \ref{tab:collision-vessel-positions} depicts this information in a concise tabular format.

\begin{table}
  \centering
  \begin{tabular}{|c|ccc|}
    \hline
    \textbf{Vessel} & \textbf{Time $t-1$} & \textbf{Time $t$} & \textbf{Time $t+1$} \\
    \hline
    $i$ & $v_{i, t-1}$ & $v_{i, t}$ & $v_{i, t+1}$ \\\hline
    $j$ & $v_{j, t-1}$ & $v_{j, t}$ & $v_{j, t+1}$ \\
    \hline
  \end{tabular}  \caption{Collision window for a collision at time $t$ and vertex $v$}
  \label{tab:collision-vessel-positions}
\end{table}

\begin{figure}
     \centering
     \begin{subfigure}[b]{0.2\textwidth}
         \centering
         \includegraphics[width=\textwidth]{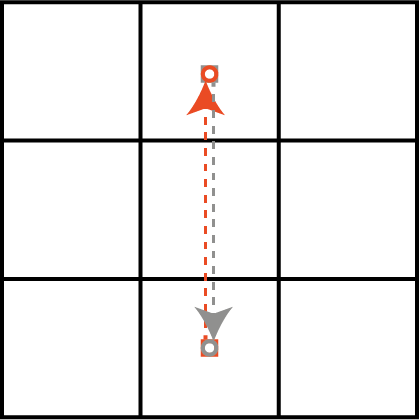}
         \caption{Head-on example}
         \label{fig:head-on-example}
     \end{subfigure}
     \begin{subfigure}[b]{0.2\textwidth}
         \centering
         \includegraphics[width=\textwidth]{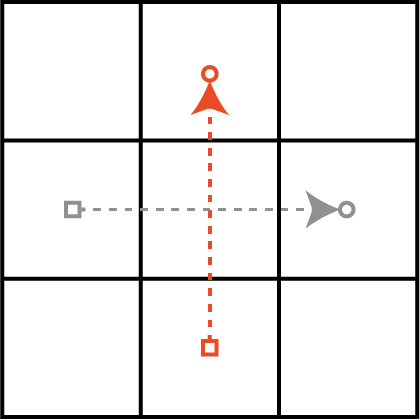}
         \caption{Crossing example}
         \label{fig:crossing-example}
     \end{subfigure}
     \\
     \begin{subfigure}[b]{0.2\textwidth}
         \centering
         \includegraphics[width=\textwidth]{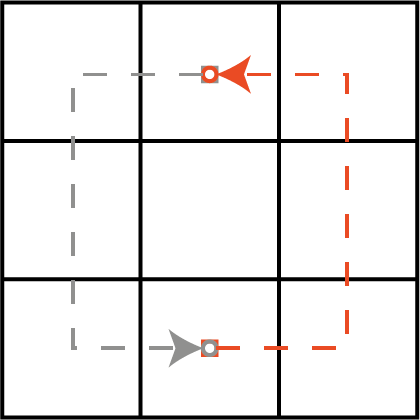}
         \caption{Head-on cell counting}
         \label{fig:head-on-counting}
     \end{subfigure}
     \begin{subfigure}[b]{0.2\textwidth}
         \centering
         \includegraphics[width=\textwidth]{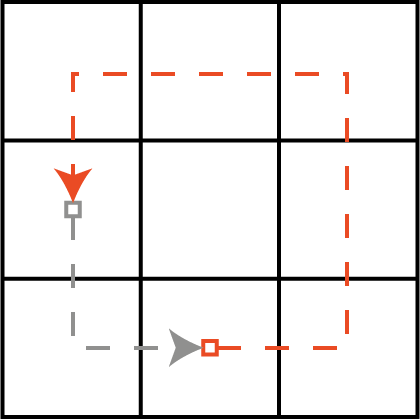}
         \caption{Crossing cell counting}
         \label{fig:crossing-counting}
     \end{subfigure}
        \caption{Example collision grids and cell counting process. Box marks vertex $v_{t-1}$ and circle marks vertex $v_{t+1}$ for both agents.}
        \label{fig:collision-grids-and-counting}
\end{figure}

The vertex of each agent at time $t-1$ in the collision window determines the type of collision situation that is created. We use a $3 \times 3$ grid to plot the trajectories of the two agents in the collision window. We call this grid the \textit{collision grid}. The collision vertex $v_{t}$ lies in the centre of the collision grid. As an agent may only travel a single vertex in one time step, vertices $v_{i, t-1}, v_{j, t-1}, v_{i, t+1} \,\&\, v_{j, t+1}$ must all lie in the 4 outer cells of the collision grid (top, left, bottom, and right). Figures \ref{fig:head-on-example} and \ref{fig:crossing-example} show an example of collision grids in the Head-on and Crossing situations respectively. To identify the conflict type, we begin counting the number of cells from the starting vertex $v_{i, t-1}$ of agent $i$ to the starting vertex $v_{j, t-1}$ of agent $j$ and vice versa. We perform this counting in anti-clockwise order along the 8 outer cells of the grid. The count value determines the conflict type in the collision. Figures \ref{fig:head-on-counting} and \ref{fig:crossing-counting} demonstrate this counting procedure, and table \ref{tab:collision-grid-counting} identifies the conflict type based on the count values.

\begin{table}
  \centering
  \begin{tabular}{|ccc|c|}
    \hline
    \textbf{Counting from} & \textbf{Counting to} & \textbf{Counts} & \textbf{Type}\\
    \hline
    $v_{i, t-1}$ & $v_{j, t-1}$ & $4$ & Head-on\\ \hline
    $v_{j, t-1}$ & $v_{i, t-1}$ & $4$ & Head-on\\ \hline
    $v_{i, t-1}$ & $v_{j, t-1}$ & $6$ & Crossing\\ \hline
    $v_{i, t-1}$ & $v_{j, t-1}$ & $2$ & Crossing\\ \hline
    $v_{j, t-1}$ & $v_{i, t-1}$ & $6$ & Crossing\\ \hline
    $v_{j, t-1}$ & $v_{i, t-1}$ & $2$ & Crossing\\
    \hline
  \end{tabular}
  \caption{Conflict type based on counting in collision grid}
  \label{tab:collision-grid-counting}
\end{table}

\subsubsection{Conflict Resolution}
The \textit{resolve()} method takes a node's solution and the identified conflict to modify the solution and avoid collisions based on the conflict information. Since it is provided only one conflict object, the \textit{resolve()} method resolves single conflicts between only two vessels at a time. The conflict resolution strategy varies based on the $ctype$ field of the conflict tuple. According to Rules 14-17 of COLREGs, the action by both vessels in Head-on situations is identical. However, in Crossing situations, the vessels must first be labeled as \textbf{Stand-on} or \textbf{Give-way} to decide collision-avoidance actions. After identifying and labeling the vessels, the \textit{resolve()} method modifies the paths in the solution in accordance with COLREGs. The modified solution is returned at the end, to be used to initialise the next node in SL.

\subsubsection{Vessel Labeling}
Since the collision avoidance actions by both vessels are identical in Head-on situations, we do not need to label the vessels as Stand-on or Give-way. However, for Crossing situations, the labeling is required. Given a conflict tuple, we again use a collision window and its collision grid of the conflict to complete the labeling. Similar to the \textit{validate()} method, we count the number of cells from the starting vertex $v_{i, t-1}$ of agent $i$ to the starting vertex $v_{j, t-1}$ of agent $j$ and vice versa in the collision grid. Table \ref{tab:labeling-by-grid-counting} demonstrates this labeling for various counts.

\begin{table}[ht]
  \centering
  \begin{tabular}{|ccc|c|c|}
    \hline
    \textbf{Counting from} & \textbf{Counting to} & \textbf{Counts} & \textbf{Agent $i$} & \textbf{Agent $j$}\\
    \hline
    $v_{i, t-1}$ & $v_{j, t-1}$ & $4$ & -        & -       \\ \hline
    $v_{j, t-1}$ & $v_{i, t-1}$ & $4$ & -        & -       \\ \hline
    $v_{i, t-1}$ & $v_{j, t-1}$ & $6$ & Stand-on & Give-way\\ \hline
    $v_{i, t-1}$ & $v_{j, t-1}$ & $2$ & Give-way & Stand-on\\ \hline
    $v_{j, t-1}$ & $v_{i, t-1}$ & $6$ & Give-way & Stand-on\\ \hline
    $v_{j, t-1}$ & $v_{i, t-1}$ & $2$ & Stand-on & Give-way\\
    \hline
  \end{tabular}
    \caption{Vessel labeling based on counting in collision grid}
  \label{tab:labeling-by-grid-counting}
\end{table}

\subsubsection{Modifying the Solution for COLREGs-Compliance}
Once the vessels have been labeled, we use this information in addition to the conflict tuple to modify the solution paths. We again make use of the collision window and collision grid. We replace the 3 vertices in the collision window for each vessel and insert a list of new vertices in their place to make the vessels' paths COLREGs-compliant. We call this list of new vertices an \textit{insertion list} (L) The modification through vertex-insertion for Head-on and Crossing collisions is given as follows:

\begin{itemize}
    \item \textbf{Head-on}: Each vessel must start from its start vertex $v_{t-1}$ and follow a sequence of cells in anti-clockwise order along the outer cells of the collision grid until it reaches its final vertex $v_{t+1}$. Algorithm \ref{alg: head-on-path-modification} gives the exact procedure to be followed to generate L.
    \item \textbf{Crossing}: The Stand-on vessel $a_{s}$ must follow the same trajectory as represented in the collision window with no modifications ($v_{s, t-1} \rightarrow v_{s, t} \rightarrow v_{s, t+1}$). Hence, we simply populate L with the same 3 vertices as in the collision window. The Give-way vessel $a_{g}$ must start from its start vertex  $v_{g, t-1}$ and follow a continuous sequence of cells along the collision grid's outer cells in anti-clockwise order till it reaches the start vertex $v_{s, t-1}$ of the Stand-on vessel $a_{s}$. If $v_{s, t-1}$ and Give-way vessel $a_{g}$'s $v_{g, t-1}$ are the same, no further insertions are needed. Otherwise, it must then visit the collision grid centre $v_{g, t}$, and then the final vertex $v_{g, t+1}$. Algorithm \ref{alg: crossing-give-way-path-modification} shows how to generate L for Give-way vessels in Crossing situations.
\end{itemize}

\begin{figure}[ht]
     \centering
     \begin{subfigure}[b]{0.2\textwidth}
         \centering
         \includegraphics[width=\textwidth]{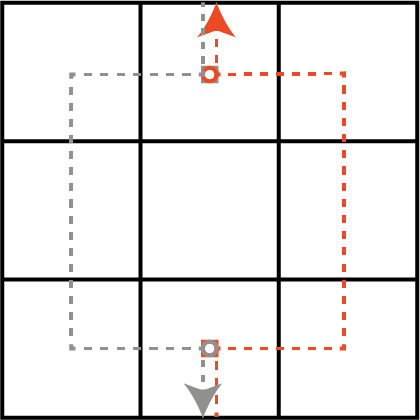}
         \caption{Safe Head-on situation}
         \label{fig:colreg-compliant-head-on}
     \end{subfigure}
     \begin{subfigure}[b]{0.2\textwidth}
         \centering
         \includegraphics[width=\textwidth]{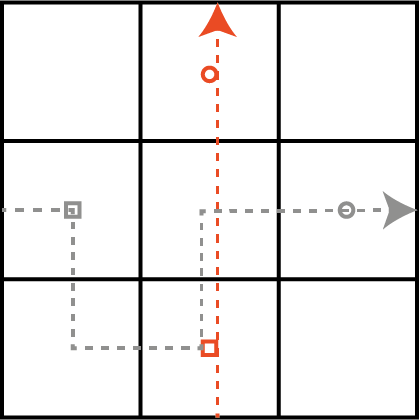}
         \caption{Safe Crossing situation}
         \label{fig:colreg-compliant-crossing}
     \end{subfigure}
        \caption{Examples of COLREGs situations resolved using COMCORE strategy.}
        \label{fig:colreg-compliant-situations}
\end{figure}

Replacing the vertices in the collision window with a list of new vertices L generated by the above process resolves the conflict and makes the involved vessels pass each other in a safe, COLREGs-compliant fashion. Figure \ref{fig:colreg-compliant-situations} shows an example implementation of this strategy in Head-on and Crossing situations.

\begin{algorithm}
\caption{Path modification for vessel in Head-on situations}
\begin{algorithmic}[1]
    \State Input: Collision window, collision grid, conflict tuple
    \State Output: Path subsection $L$
    \State $L\gets$ \textsc{append}($L,v_{t-1}$) \Comment{Append the starting vertex}
    \While{$L[-1] \neq v_{t+1}$}
        \State $v_n \gets$ next outer cell in anti-clockwise order
        \State $L\gets$ \textsc{append}($L,v_n$)
    \EndWhile
\end{algorithmic}
\label{alg: head-on-path-modification}
\end{algorithm}

\begin{algorithm}
\caption{Path modification for Give-way vessel in Crossing situations}
\begin{algorithmic}[1]
    \State Input: Collision window, collision grid, conflict tuple
    \State Output: Path subsection $L$
    \State $L\gets$ \textsc{append}($L,v_{g, t-1}$) \Comment{Append the starting vertex}
    \While{$L[-1] \neq v_{s, t-1}$}
        \State $v_n \gets$ next outer cell in anti-clockwise order
        \State $L\gets$ \textsc{append}($L,v_n$)
    \EndWhile
    \If{$L[-1] \neq v_{g, t+1}$} \Comment{If not the final vertex}
        \State $L\gets$ \textsc{append}($L,v_{g, t}$) \Comment{Insert the central vertex}
        \State $L\gets$ \textsc{append}($L,v_{g, t+1}$) \Comment{Insert the final vertex}
    \EndIf
\end{algorithmic}
\label{alg: crossing-give-way-path-modification}
\end{algorithm}

\section{EXPERIMENTAL RESULTS}
In this section, we demonstrate the validity of the COMCORE strategy through simulations and field experiments. 
\subsection{Simulation setup}
We consider a workspace of 70m $\times$ 70m. The workspace is discretized into cells of 10m length and width respectively. Thus, the workspace is composed of a 7 $\times$ 7 grid. We consider two agents and different collision conditions to show the efficacy of the COMCORE to resolve conflicts while satisfying the COLREGs conditions.  We assign the start and goal position for each vessel in the 7$\times$ 7 grid.  The speed of the vehicle is approximately 2 knots. The agents initiate the COMCORE strategy when they are 3 cells apart. 

\subsection{Simulation Results}
\begin{figure}
     \centering
     \begin{subfigure}[b]{0.2\textwidth}
         \centering
         \includegraphics[width=\textwidth]{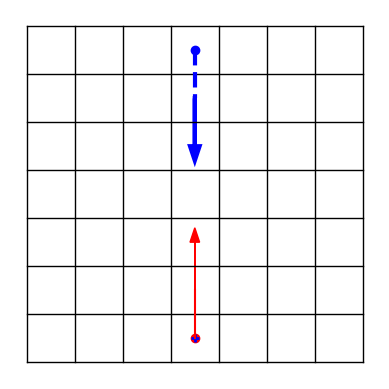}
         \caption{}
     \end{subfigure}
     \begin{subfigure}[b]{0.2\textwidth}
         \centering
         \includegraphics[width=\textwidth]{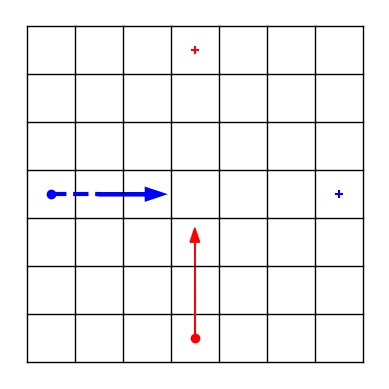}
         \caption{}
     \end{subfigure}
        \caption{Scenarios for (a) Head-on collision and (b) Crossing. $\bullet$ is the start location of the vehicle. $\rightarrow$ represents the direction of travel and current location of the vessel. {\bf{+}} is the destination of the grid of the vessel.}
        \label{fig:sim_scenarios}
\end{figure}

\begin{figure}
     \centering
     \begin{subfigure}[b]{0.2\textwidth}
         \centering
         \includegraphics[width=\textwidth]{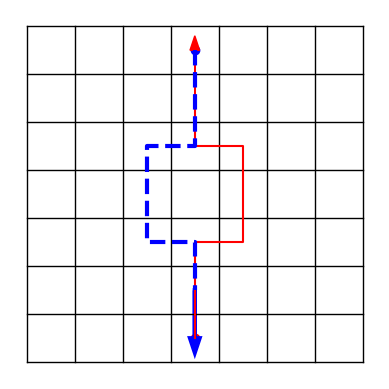}
         \caption{}
         \label{fig:result-4b-simulation}
     \end{subfigure}
     \begin{subfigure}[b]{0.2\textwidth}
         \centering
         \includegraphics[width=\textwidth]{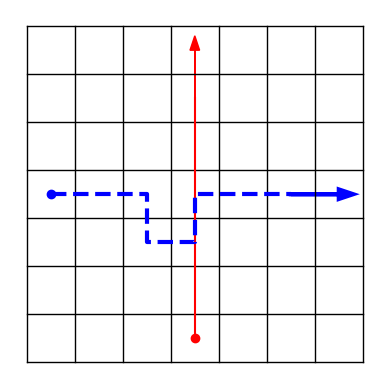}
         \caption{}
         \label{fig:result-1a-simulation}
     \end{subfigure}
        \caption{COMCORE-resolved (a)  Head-on situation (b) Crossing situation }
        \label{fig:result-typical-collisions}
\end{figure}

In this section, we demonstrate the COMCORE strategy on two common conflict scenarios -- (i) head-on collision and (ii) crossing scenario (shown in Figure \ref{fig:sim_scenarios}). The two agents (red and blue) detect the conflict and execute the COMCORE strategy independently. The resultant solution generated by the COMCORE strategy is shown in Figure \ref{fig:result-typical-collisions} for both the scenarios. From the figure, we can see the COMCORE strategy resolves the conflict with minimal deviation. 

Consider the head-on collision scenario as shown in Figure \ref{fig:sim_scenarios}(a), where the agents are initialized at "$\bullet$" and travel towards their respective goal locations given by "{\bf +}". The current location of the agents in given by "$\rightarrow$". If the agents navigate along their path towards the goal without any collision avoidance actions, they would would collide at time step $t=3s$, in the middle of the workspace. However, at time step $t=2$, COMCORE is initiated, which provides a conflict-free path for each vessel as shown in Figure \ref{fig:result-typical-collisions}(a). It is also important to note that the action by the Stand-on and Give-way vessels differs in the Crossing situation, with the Stand-on vessel maintaining its course and the Give-way vessel manoeuvring around it.

\begin{figure}
     \centering
     \begin{subfigure}[b]{0.2\textwidth}
         \centering
         \includegraphics[width=\textwidth]{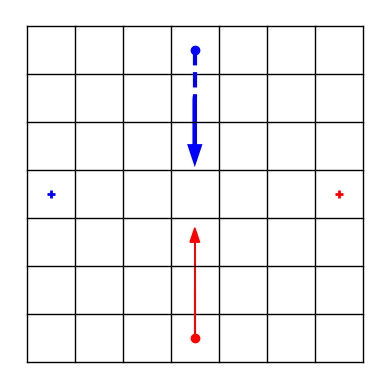}        
         \caption{}
         \label{fig:result-1d-scenario-simulation}
     \end{subfigure}
     \begin{subfigure}[b]{0.2\textwidth}
         \centering
         \includegraphics[width=\textwidth]{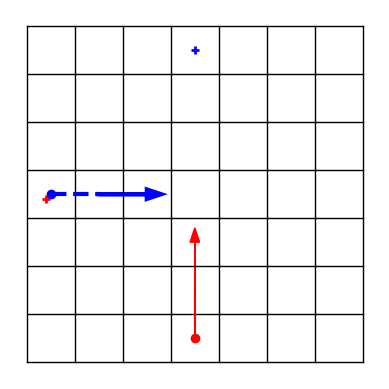}
         \caption{}
         \label{fig:result-2c-scenario-simulation}
     \end{subfigure}
        \caption{Some more COLREGs scenarios for (a) Head-on and (b) Crossing situations.}
        \label{fig:additional-simulation-scenario-collisions}
\end{figure}

\begin{figure}[ht]
     \centering
     \begin{subfigure}[b]{0.2\textwidth}
         \centering
         \includegraphics[width=\textwidth]{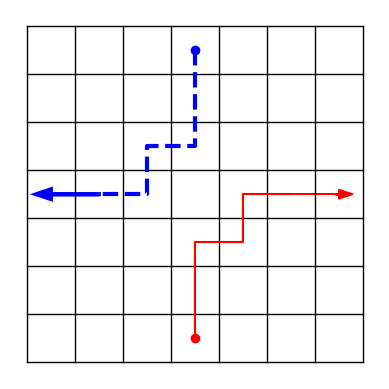}
         \caption{}
         \label{fig:result-1d-simulation}
     \end{subfigure}
     \begin{subfigure}[b]{0.2\textwidth}
         \centering
         \includegraphics[width=\textwidth]{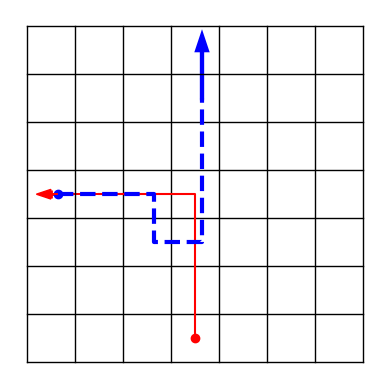}
         \caption{}
         \label{fig:result-2c-simulation}
     \end{subfigure}
        \caption{Some more COLREGs-compliant paths for (a) Head-on and (b) Crossing situations.}
        \label{fig:additional-simulation-collisions}
\end{figure}

Additional scenarios for the head-on and crossing collision scenarios are shown in Figure \ref{fig:additional-simulation-scenario-collisions}. In Figure \ref{fig:additional-simulation-scenario-collisions}(a), the two vessels may collide one step away, but with COMCORE the resolved trajectories are shown in Figure \ref{fig:additional-simulation-collisions}(a). We can see that the vessels take evasive action quickly and resolve the conflict. As similar scenario is shown in Figure \ref{fig:additional-simulation-scenario-collisions}(b) for the crossing situation. In this case, the blue vessel takes a de-tour while the red vessel continues in its pre-assigned trajectory. Even in this case, we can see that the deviation from the assigned trajectory towards the goal is  minimal. 

\begin{figure}
    \centering
    \includegraphics[width=\linewidth]{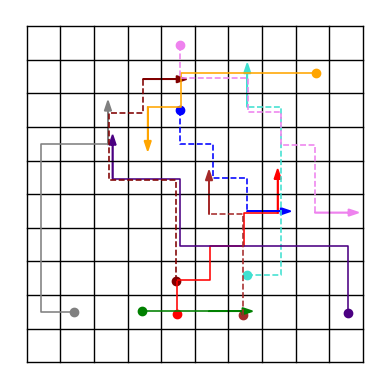}
    \caption{Multi-agent COLREGs-compliant paths using COMCORE.}
    \label{fig:multi-agent-paths}
\end{figure}

We now extend the simulation to 10 agents as shown in Figure \ref{fig:multi-agent-paths}. The resultant paths clearly show that ability of the COMCORE strategy to resolve conflicts and scale. 






\begin{figure}
    \centering
    \includegraphics[width=\linewidth]{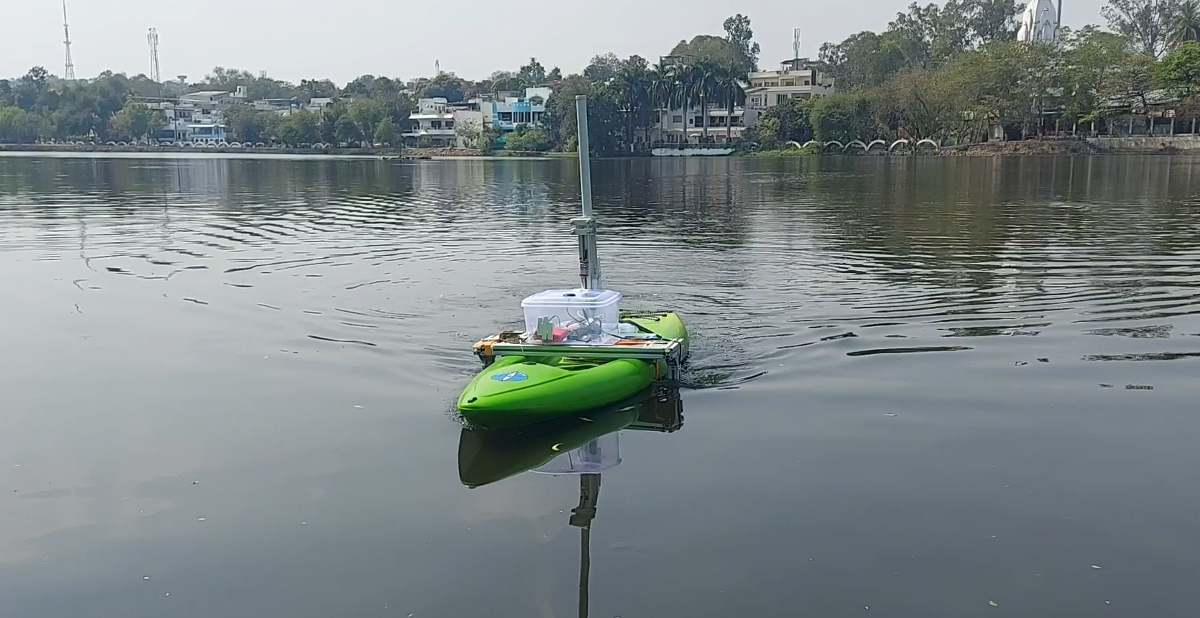}
    \caption{ASV ``Dolphin" at Lower Lake, Bhopal, India.}
    \label{fig:dolphin-on-lake}
\end{figure}

\subsection{Physical Experimental Results}
We evaluate the complete functionality of the COMCORE strategy  through real-world demonstrations on an ASV. The experiment was carried out in the Bhopal Lower Lake, Bhopal, India. We have developed an in-house ASV based on a single hull kayak hull (called ``Dolphin"). The Figure \ref{fig:dolphin-on-lake} shows the vessel navigating in the water autonomously. 

\subsubsection{Dolphin vessel} The vessel is a single hull kayak of length 2400mm made from plastic. The width of the vessel is 700mm and hull weight is 13Kg with a payload capacity of 75Kg. The vehicle is equipped with T200 thrusters that create a differential drive mechanism. It has a Pixhawk 2.1 Cube autopilot with ArduRover firmware. We built an external frame using aluminium extrusions weighing about 10Kg to mount the various components on the vessel. These components also add approximately 8Kg to the payload. The onboard computer is a Jetson Nano. For communication to the base station, RFD 868 radio modems are used and a Ubiquity Rocket M5 with an onmi-directional antenna is used for communication with other vessels. The maximum speed of Dolphin is 4 knots.

The ASV broadcasts its global location to all nearby vessels. When two independently-navigating vessels come closer than a threshold distance, they switch to planning using COMCORE and computing a collision-free path to their target. Currently, due to lack of time, we were able to build only one ASV and hence we demonstrate the COMCORE strategy on Dolphin while we simulate the other vehicle's trajectory in real-time on-board the vessel. 
The resolved trajectories enable us to validate the performance of COMCORE for generating COLREGs-compliant manoeuvres. 

\subsubsection{Experiment}
\begin{figure}[ht]
     \centering
     \begin{subfigure}[b]{0.225\textwidth}
         \centering
         \includegraphics[width=\textwidth]{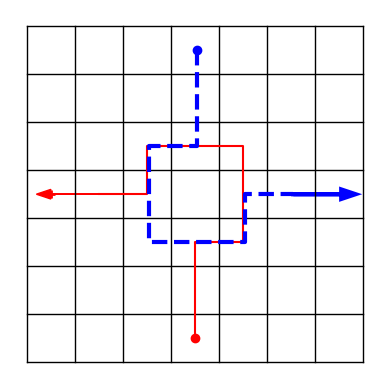}
         \caption{}
         \label{fig:result-5b-simulation}
     \end{subfigure}
     \begin{subfigure}[b]{0.225\textwidth}
         \centering
         \includegraphics[width=\textwidth]{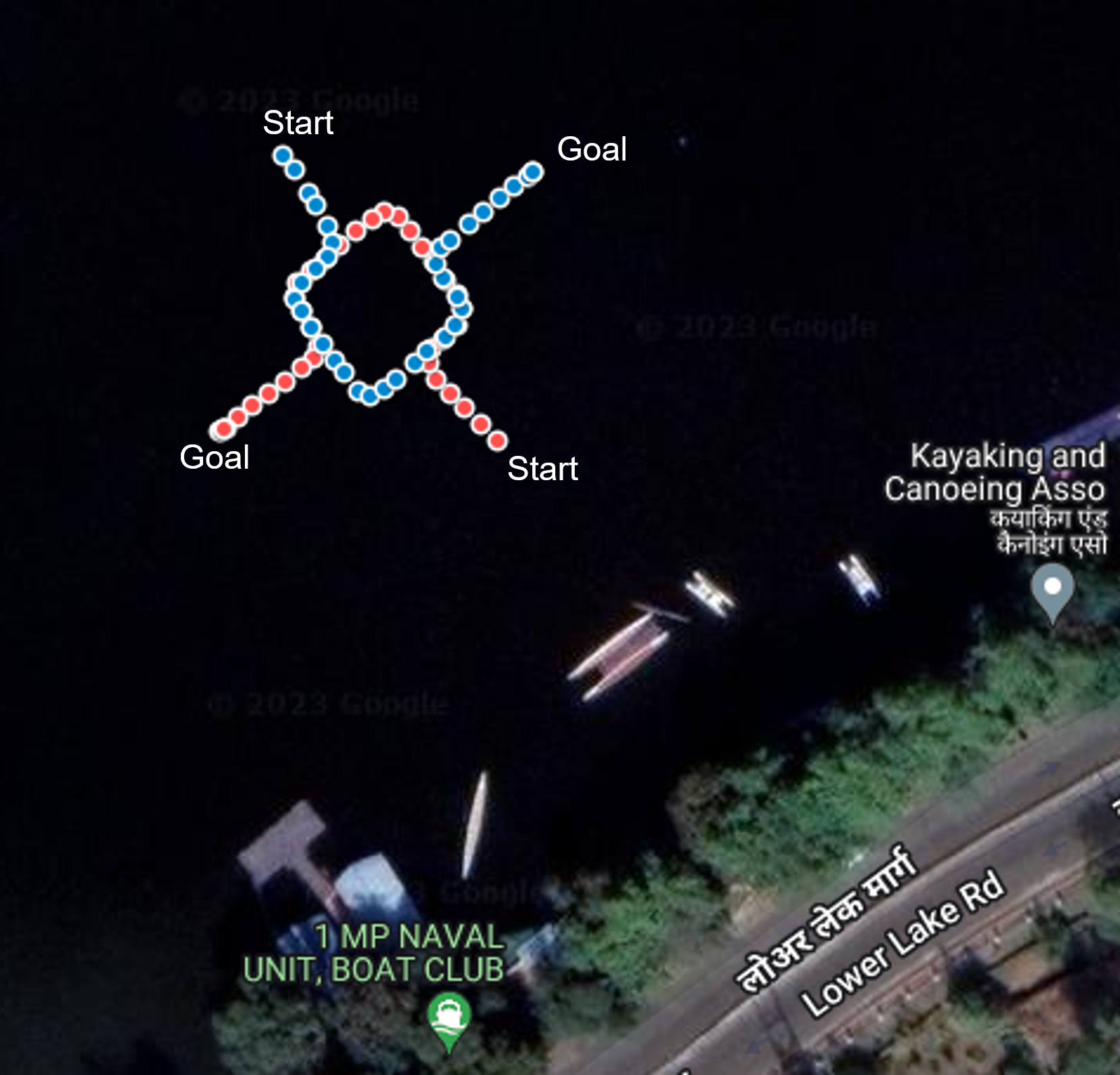}
         \caption{}
         \label{fig:result-5b-gps}
     \end{subfigure}
     \\
     \begin{subfigure}[b]{0.225\textwidth}
         \centering
         \includegraphics[width=\textwidth]{Diagrams/simulation_results/2c.png}
         \caption{}
         \label{fig:result-2c-simulation1}
     \end{subfigure}
     \begin{subfigure}[b]{0.225\textwidth}
         \centering
         \includegraphics[width=\textwidth]{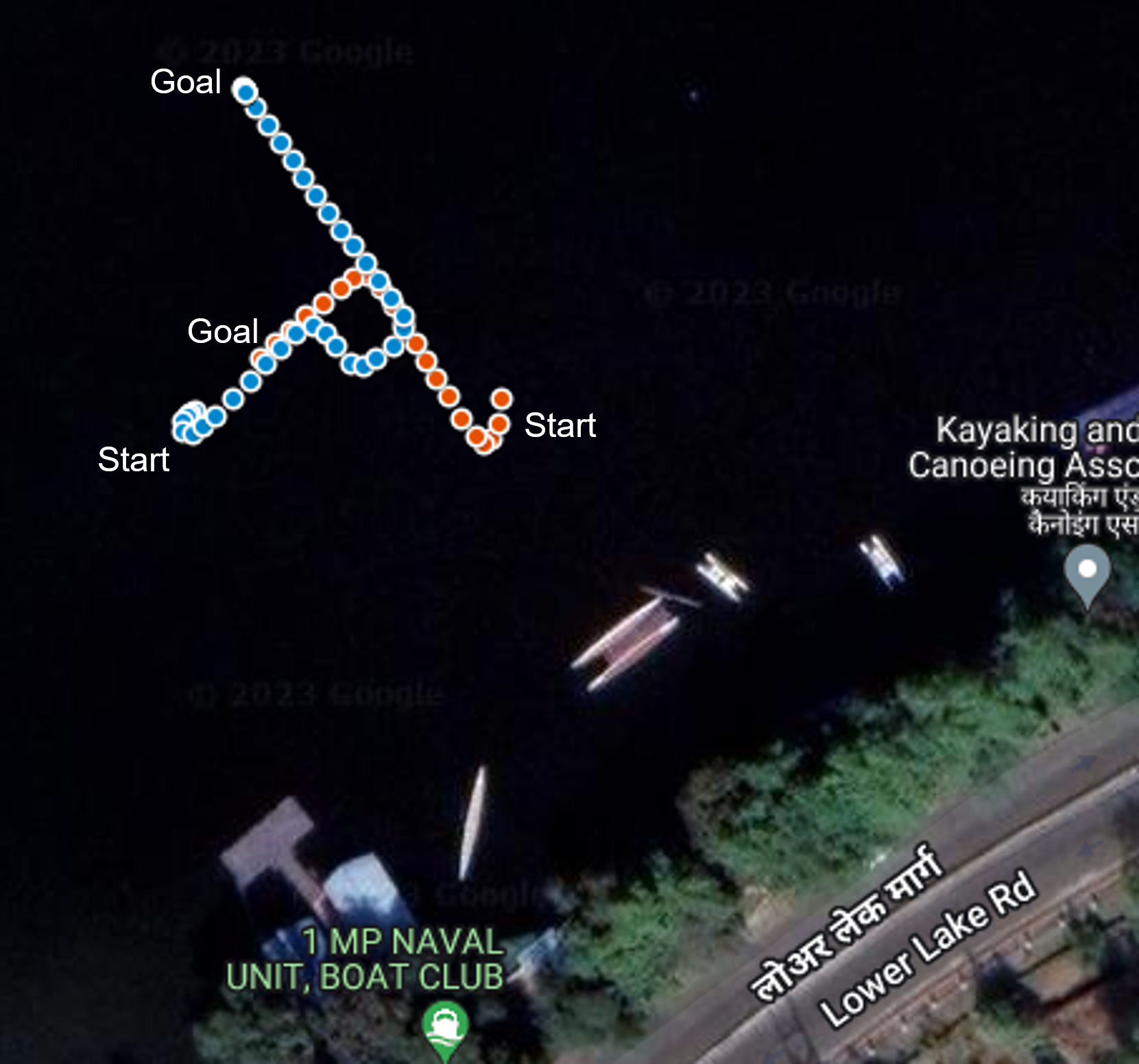}
         \caption{}
         \label{fig:result-2c-gps}
     \end{subfigure}
        \caption{Computed paths in a discrete workspace vs actual experimental ASV trajectories.}
        \label{fig:simulation-vs-gps-paths}
\end{figure}

We consider an area of $70m \times 70m$ in the lake. The area is discretized into 7 $\times$ 7 cells, each cell having a 10m width and length. Initially, the vessels are located at the start location and move towards the goal location. When the vessels are 30m (3 cells) apart, they check for potential collisions and the COMCORE algorithm is invoked by Dolphin and the simulated vessel simultaneously. The resultant generated solution is shown in Figure  \ref{fig:simulation-vs-gps-paths}. The conflict-avoidance actions follow the prescribed actions in COLREGs. Figure \ref{fig:simulation-vs-gps-paths}(a) is a tricky Head-on situation, and the vessels each take action. The vessels consider the COMCORE solution as a set of waypoints and follow this new plan executing the conflict resolution. Figure \ref{fig:simulation-vs-gps-paths}(b) shows the GPS trajectories of the two vessels. The trajectories shown in the Figure are the actual GPS trajectories. 

Further, we demonstrate the crossing scenario using the ASV as shown in Figure \ref{fig:simulation-vs-gps-paths}(c) and (d). 

From the simulations and real-world experiments, we have demonstrated that the COMCORE strategy can successfully generate conflict free paths satisfying COLREG requirements for ASVs. 

\section{CONCLUSIONS}
In this paper, we have presented a path planning strategy for ASVs using a conflict-resolution approach. We have provided details of the algorithms and demonstrated its capabilities through simulation and real-world experiments on an ASV. Our simulation and experimental tests validated the utility of the COMCORE strategy in real-life scenarios, and provided insights into the application of discrete conflict resolution algorithms in practical situations.

 The present work can be extended in terms of determining the theoretical properties of COMCORE and extend towards analyzing the effect of communication lags while determining the solution. Another extended version of the problem can consider the currents and winds while determining the paths. The efficacy of the proposed approach against existing approaches needs to be analyzed. 

\bibliographystyle{IEEEtran}
\bibliography{references}

\begin{thebibliography}{10}
\providecommand{\url}[1]{#1}
\csname url@samestyle\endcsname
\providecommand{\newblock}{\relax}
\providecommand{\bibinfo}[2]{#2}
\providecommand{\BIBentrySTDinterwordspacing}{\spaceskip=0pt\relax}
\providecommand{\BIBentryALTinterwordstretchfactor}{4}
\providecommand{\BIBentryALTinterwordspacing}{\spaceskip=\fontdimen2\font plus
\BIBentryALTinterwordstretchfactor\fontdimen3\font minus
  \fontdimen4\font\relax}
\providecommand{\BIBforeignlanguage}[2]{{%
\expandafter\ifx\csname l@#1\endcsname\relax
\typeout{** WARNING: IEEEtran.bst: No hyphenation pattern has been}%
\typeout{** loaded for the language `#1'. Using the pattern for}%
\typeout{** the default language instead.}%
\else
\language=\csname l@#1\endcsname
\fi
#2}}
\providecommand{\BIBdecl}{\relax}
\BIBdecl

\bibitem{van2011reciprocal}
J.~Van Den~Berg, J.~Snape, S.~J. Guy, and D.~Manocha, ``Reciprocal collision
  avoidance with acceleration-velocity obstacles,'' in \emph{2011 IEEE
  International Conference on Robotics and Automation}.\hskip 1em plus 0.5em
  minus 0.4em\relax IEEE, 2011, pp. 3475--3482.

\bibitem{fiorini1998motion}
P.~Fiorini and Z.~Shiller, ``Motion planning in dynamic environments using
  velocity obstacles,'' \emph{The international journal of robotics research},
  vol.~17, no.~7, pp. 760--772, 1998.

\bibitem{khatib1986real}
O.~Khatib, ``Real-time obstacle avoidance for manipulators and mobile robots,''
  \emph{The international journal of robotics research}, vol.~5, no.~1, pp.
  90--98, 1986.

\bibitem{sun2017collision}
J.~Sun, J.~Tang, and S.~Lao, ``Collision avoidance for cooperative uavs with
  optimized artificial potential field algorithm,'' \emph{IEEE Access}, vol.~5,
  pp. 18\,382--18\,390, 2017.

\bibitem{roussos20083d}
G.~P. Roussos, D.~V. Dimarogonas, and K.~J. Kyriakopoulos, ``3d navigation and
  collision avoidance for a non-holonomic vehicle,'' in \emph{2008 American
  Control Conference}.\hskip 1em plus 0.5em minus 0.4em\relax IEEE, 2008, pp.
  3512--3517.

\bibitem{panagou2014motion}
D.~Panagou, ``Motion planning and collision avoidance using navigation vector
  fields,'' in \emph{2014 IEEE International Conference on Robotics and
  Automation (ICRA)}.\hskip 1em plus 0.5em minus 0.4em\relax IEEE, 2014, pp.
  2513--2518.

\bibitem{seiler1998application}
P.~Seiler, B.-S. Song, and J.~K. Hedrick, ``Application of nonlinear control to
  a collision avoidance system,'' \emph{Proc. 5th World Congr. ITS}, pp.
  12--16, 1998.

\bibitem{hu2020real}
Y.~Hu, X.~Meng, Q.~Zhang, and G.-K. Park, ``A real-time collision avoidance
  system for autonomous surface vessel using fuzzy logic,'' \emph{Ieee Access},
  vol.~8, pp. 108\,835--108\,846, 2020.

\bibitem{chakravarthy1998obstacle}
A.~Chakravarthy and D.~Ghose, ``Obstacle avoidance in a dynamic environment: A
  collision cone approach,'' \emph{IEEE Transactions on Systems, Man, and
  Cybernetics-Part A: Systems and Humans}, vol.~28, no.~5, pp. 562--574, 1998.

\bibitem{zhang21}
X.~Zhang, C.~Wang, L.~Jiang, L.~An, and R.~Yang, ``Collision-avoidance
  navigation systems for maritime autonomous surface ships: A state of the art
  survey,'' \emph{Ocean Engineering}, vol. 235, p. 109380, 2021.

\bibitem{8281087}
H.-T.~L. Chiang and L.~Tapia, ``Colreg-rrt: An rrt-based colregs-compliant
  motion planner for surface vehicle navigation,'' \emph{IEEE Robotics and
  Automation Letters}, vol.~3, no.~3, pp. 2024--2031, July 2018.

\bibitem{8463182}
I.~B. Hagen, D.~K.~M. Kufoalor, E.~F. Brekke, and T.~A. Johansen, ``Mpc-based
  collision avoidance strategy for existing marine vessel guidance systems,''
  in \emph{2018 IEEE International Conference on Robotics and Automation
  (ICRA)}, May 2018, pp. 7618--7623.

\bibitem{kufoalor2018proactive}
D.~K.~M. Kufoalor, E.~F. Brekke, and T.~A. Johansen, ``Proactive collision
  avoidance for asvs using a dynamic reciprocal velocity obstacles method,'' in
  \emph{2018 IEEE/RSJ International Conference on Intelligent Robots and
  Systems (IROS)}.\hskip 1em plus 0.5em minus 0.4em\relax IEEE, 2018, pp.
  2402--2409.

\bibitem{thyri2022partly}
E.~H. Thyri and M.~Breivik, ``Partly colregs-compliant collision avoidance for
  asvs using encounter-specific velocity obstacles,'' \emph{IFAC-PapersOnLine},
  vol.~55, no.~31, pp. 37--43, 2022.

\bibitem{SHARON201540}
\BIBentryALTinterwordspacing
G.~Sharon, R.~Stern, A.~Felner, and N.~R. Sturtevant, ``Conflict-based search
  for optimal multi-agent pathfinding,'' \emph{Artificial Intelligence}, vol.
  219, pp. 40--66, 2015. [Online]. Available:
  \url{https://www.sciencedirect.com/science/article/pii/S0004370214001386}
\BIBentrySTDinterwordspacing

\bibitem{4082128}
P.~E. Hart, N.~J. Nilsson, and B.~Raphael, ``A formal basis for the heuristic
  determination of minimum cost paths,'' \emph{IEEE Transactions on Systems
  Science and Cybernetics}, vol.~4, no.~2, pp. 100--107, July 1968.

\end{thebibliography}

\end{document}